\documentclass[11pt,a4paper]{article}
\usepackage[hyperref]{emnlp-ijcnlp-2019}
\usepackage{times}
\usepackage{latexsym}
\usepackage{graphicx}
\usepackage{amsmath}
\usepackage{multirow}
\usepackage{array}

\usepackage{url}

\aclfinalcopy 

\title{Table-to-Text Generation with Effective Hierarchical Encoder \\ on Three Dimensions (Row, Column and Time)}

\author{Heng Gong, Xiaocheng Feng, Bing Qin\thanks{\ \ \  Email corresponding.}, Ting Liu \\Harbin Institute of Technology, China \\{\tt \{hgong, xcfeng, qinb, tliu\}@ir.hit.edu.cn} }

\date{}

\begin{document}
\maketitle
\begin{abstract}
Although Seq2Seq models for table-to-text generation have achieved remarkable progress, modeling table representation in one dimension is inadequate.
This is because (1) the table consists of multiple rows and columns, which means that encoding a table should not depend only on one dimensional sequence or set of records and (2) most of the tables are time series data (e.g. NBA game data, stock market data), which means that the description of the current table may be affected by its historical data.
To address aforementioned problems, not only do we model each table cell considering other records in the same row, we also enrich table's representation by modeling each table cell in context of other cells in the same column or with historical (time dimension) data respectively. In addition, we develop a table cell fusion gate to combine representations from row, column and time dimension into one dense vector according to the saliency of each dimension's representation.
We evaluated our methods on ROTOWIRE, a benchmark dataset of NBA basketball games. 
Both automatic and human evaluation results demonstrate the effectiveness of our model with improvement of 2.66 in BLEU over the strong baseline and outperformance of state-of-the-art model.
\end{abstract}

\section{Introduction}
\label{intro}

Table-to-text generation is an important and challenging task in natural language processing, which aims to produce the summarization of numerical table \cite{reiter2000building,gkatzia2016content}.
The related methods can be empirically divided into two categories, pipeline model and end-to-end model.
The former consists of content selection, document planning and realisation, mainly for early industrial applications, such as weather forecasting and medical monitoring, etc.
The latter generates text directly from the table through a standard neural encoder-decoder framework to avoid error propagation and has achieved remarkable progress.
In this paper, we particularly focus on exploring how to improve the performance of neural methods on table-to-text generation.

\begin{figure}[t]
   \centering
   \begin{center}
     \includegraphics*[width=1.0\linewidth]{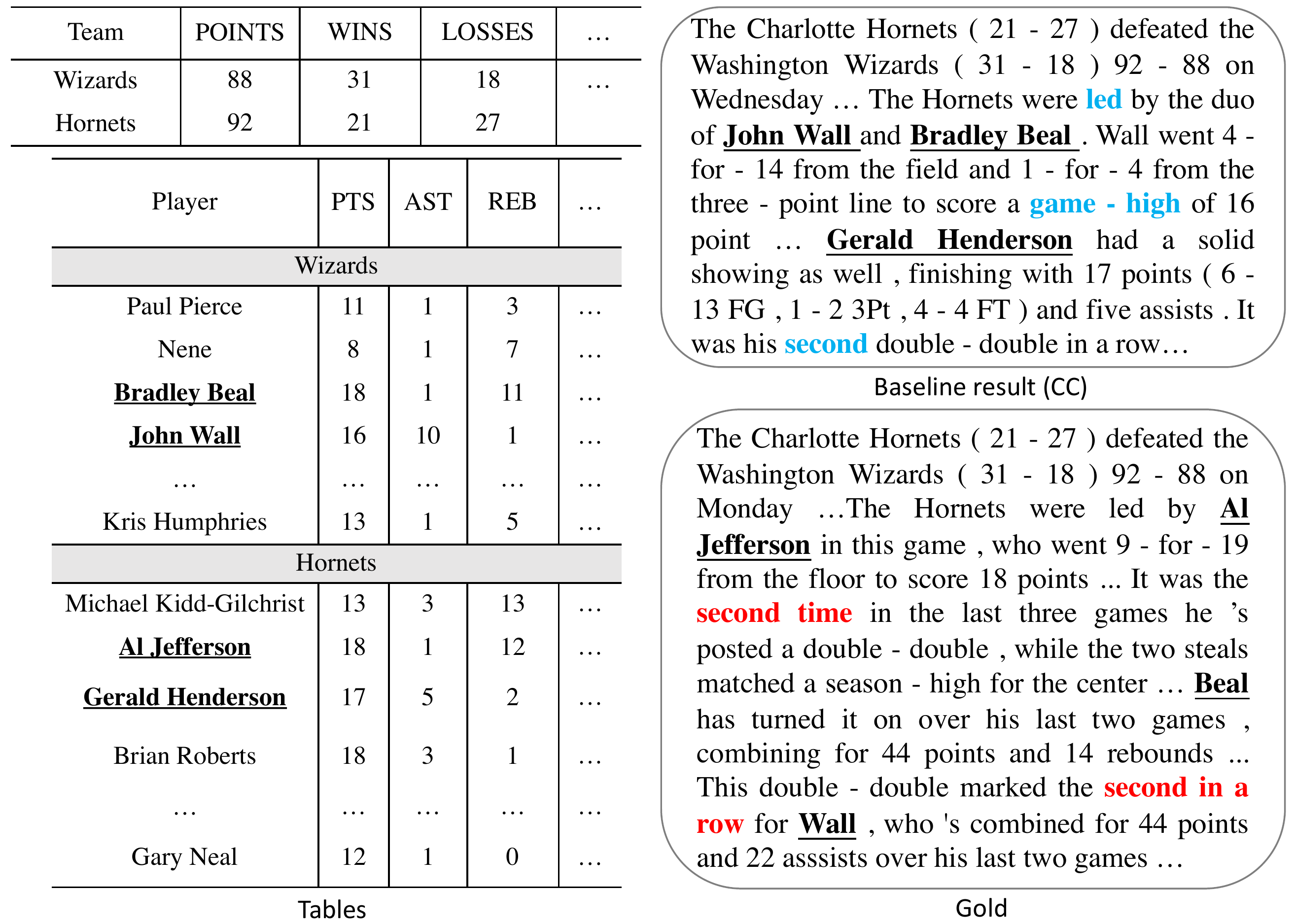}
   \caption{Generated example on ROTOWIRE by using Conditional Copy (CC) as baseline \cite{Wiseman}. Text that accurately reflects records in the table is in red, and text that contradicts the records is in blue.}
   \label{baseline-result}
   \vspace{-0.4cm}
   \end{center}
\end{figure}

Recently, ROTOWIRE, which provides tables of NBA players' and teams' statistics with a descriptive summary, has drawn increasing attention from academic community.
Figure \ref{baseline-result} shows an example of parts of a game's statistics and its corresponding computer generated summary.
We can see that the tables has a formal structure including table row header, table column header and table cells. 
``Al Jefferson'' is a table row header that represents a player, ``PTS'' is a table column header indicating the column contains player's score and ``18'' is the value of the table cell, that is, Al Jefferson scored 18 points. 
Several related models have been proposed .
They typically encode the table's records separately or as a long sequence and generate a long descriptive summary by a standard Seq2Seq decoder with some modifications.
\newcite{Wiseman} explored two types of copy mechanism and found conditional copy model \cite{Glehre2016PointingTU} perform better .
\newcite{Puduppully} enhanced content selection ability by explicitly selecting and planning relevant records. 
\newcite{Li} improved the precision of describing data records in the generated texts by generating a template at first and filling in slots via copy mechanism. 
\newcite{Nie} utilized results from pre-executed operations to improve the fidelity of generated texts.
However, we claim that their encoding of tables as sets of records or a long sequence is not suitable. Because 
(1) the table consists of multiple players and different types of information as shown in Figure \ref{baseline-result}.
The earlier encoding approaches only considered the table as sets of records or one dimensional sequence, which would lose the information of other (column) dimension. 
(2) the table cell consists of time-series data which change over time. 
That is to say, sometimes historical data can help the model select content.
Moreover, when a human writes a basketball report, he will not only focus on the players' outstanding performance in the current match, but also summarize players' performance in recent matches.
Lets take Figure \ref{baseline-result} again. Not only do the gold texts mention Al Jefferson's great performance in this match, it also states that ``It was the second time in the last three games he's posted a double-double''. Also gold texts summarize John Wall's ``double-double'' performance in the similar way. Summarizing a player's performance in recent matches requires the modeling of table cell with respect to its historical data (time dimension) which is absent in baseline model. Although baseline model Conditional Copy (CC) tries to summarize it for Gerald Henderson, it clearly produce wrong statements since he didn't get ``double-double'' in this match.

To address the aforementioned problems, 
we present a hierarchical encoder to simultaneously model row, column and time dimension information.
In detail, our model is divided into three layers. 
The first layer is used to learn the representation of the table cell.
Specifically, we employ three self-attention models to obtain three representations of the table cell in its row, column and time dimension. Then, in the second layer, we design a record fusion gate to identify the more important representation from those three dimension and combine them into a dense vector.
In the third layer, we use mean pooling method to merge the previously obtained table cell representations in the same row into the representation of the table's row. Then, we use self-attention with content selection gate \cite{Puduppully} to filter unimportant rows' information.
To the best of our knowledge, this is the first work on neural table-to-text generation via modeling column and time dimension information so far.
We conducted experiments on ROTOWIRE. 
Results show that our model outperforms existing systems, improving baseline BLEU from 14.19 to 16.85 ($+18.75\%$), P\% of relation generation (RG) from 74.80 to 91.46 ($+22.27\%$), F1\% of content selection (CS) from 32.49 to 41.21 ($+26.84\%$) and content ordering (CO) from 15.42 to 20.86 ($+35.28\%$) on test set. It also exceeds the state-of-the-art model in terms of those metrics.
\section{Preliminaries}

\subsection{Notations}
\label{task-notations}

The input to the model are tables $S=\{s^{1}, s^{2}, s^{3}\}$. $s^{1}$, $s^{2}$, and $s^{3}$ contain records about players' performance in home team, players' performance in visiting team and team's overall performance respectively. We regard each cell in the table as record. Each record $r$ consists of four types of information including value $r.v$ (e.g. 18), entity $r.e$ (e.g. Al Jefferson), type $r.c$ (e.g. POINTS) and a feature $r.f$ (e.g. visiting) which indicate whether a player or a team compete in home court or not. 
Each player or team takes one row in the table and each column contains a type of record such as points, assists, etc. Also, tables contain the date when the match happened and we let $k$ denote the date of the record. We also create timelines for records. 
The details of timeline construction is described in Section \ref{timeline-const}. For simplicity, we omit table id $l$ and record date $k$ in the following sections and let $r_{i,j}$ denotes a record of $i^{th}$ row and $j^{th}$ column in the table. We assume the records come from the same table and $k$ is the date of the mentioned record.
Given those information, the model is expected to generate text $y=(y_{1}, ..., y_{t}, ..., y_{T})$ describing these tables. $T$ denotes the length of the text.

\begin{figure*}[ht]
	\centering
	\includegraphics[width=1.0\textwidth]{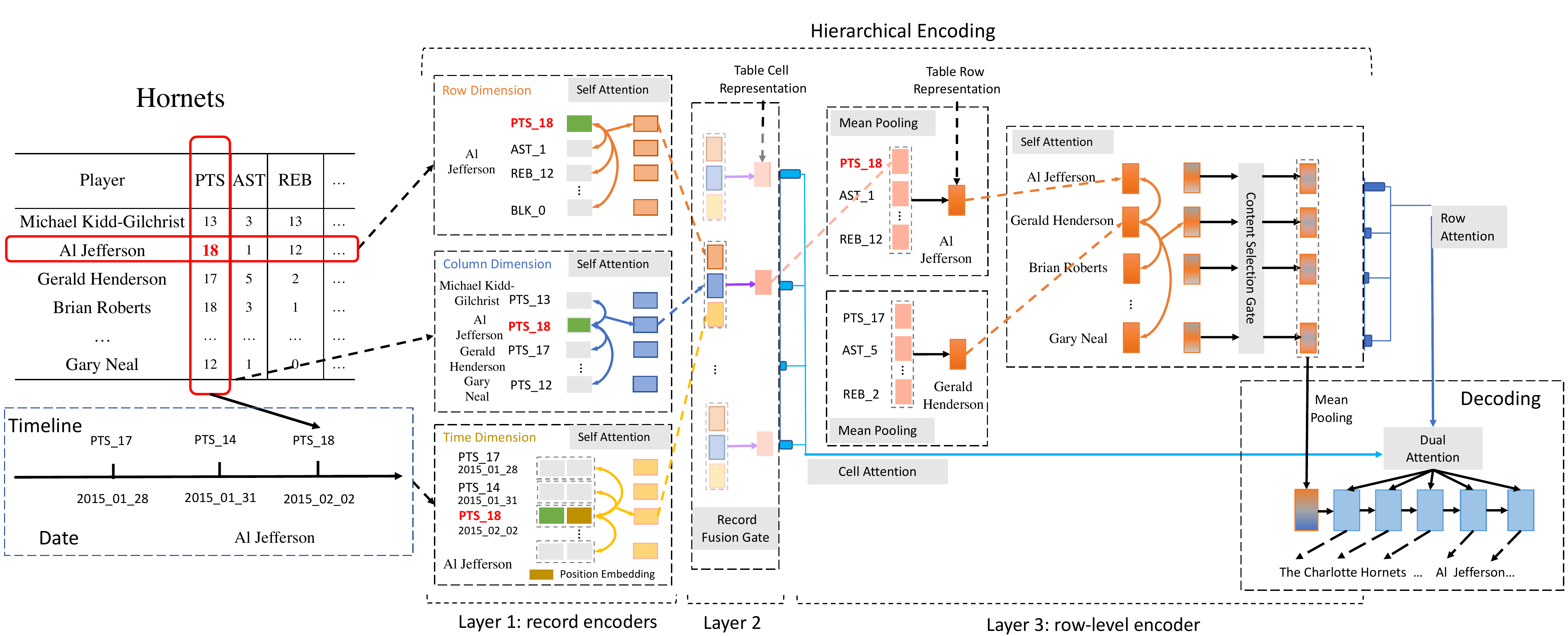}
    \vspace{-0.5cm}
            \caption{The architecture of our proposed model.}
          \label{modelstructure} 
          \vspace{-0.5cm}
\end{figure*}

\subsection{Record Timeline Constrcution}
\label{timeline-const}

In this paper, we construct timelines $tl=\{tl_{e,c}\}_{e=1,c=1}^{E,C}$ for records. $E$ denotes the number of distinct record entities and $C$ denotes the number of record types. For each timeline $tl_{e,c}$, we first extract records with the same entity $e$ and type $c$ from dataset. Then we sort them into a sequence according to the record's date from old to new. This sequence is considered as timeline $tl_{e,c}$. For example, in Figure \ref{modelstructure}, the ``Timeline'' part in the lower-left corner represents a timeline for entity Al Jefferson and type PTS (points).

\subsection{Baseline Model}
\label{base-model}

We use Seq2Seq model with attention \cite{D15-1166} and conditional copy \cite{Glehre2016PointingTU} as the base model. During training, given tables $S$ and their corresponding reference texts $y$, the model maximized the conditional probability $P(y|S) = \prod_{t=1}^{T}P(y_{t}|y_{<t},S)$ . $t$ is the timestep of decoder.
First, for each record of the $i^{th}$ row and $j^{th}$ column in the table, we utilize 1-layer MLP to encode the embeddings of each record's four types of information into a dense vector $\boldsymbol{{r}_{i, j}}$, 
$\boldsymbol{r_{i, j}}=ReLU(\boldsymbol{W_{a}}[\boldsymbol{r_{i, j}.e};\boldsymbol{r_{i, j}.c};\boldsymbol{r_{i, j}.v};\boldsymbol{r_{i, j}.f}]+\boldsymbol{b_{a}})$.
$\boldsymbol{W_{a}}$ and $\boldsymbol{b_{a}}$ are trainable parameters. The word embeddings for each type of information are trainable and randomly initialized before training following \newcite{Wiseman}. $[;]$ denotes the vector concatenation. 
Then, we use a LSTM decoder with attention and conditional copy to model the conditional probability $P(y_{t}|y_{<t},S)$.
The base model first use attention mechanism \cite{D15-1166} to find relevant records from the input tables and represent them as context vector. Please note that the base model doesn't utilize the structure of three tables and 
normalize the attention weight $\alpha_{t, i^{\prime}, j^{\prime}}$ across every records in every tables. Then it combines the context vector with decoder's hidden state $\boldsymbol{d_{t}}$ and form a new attentional hidden state $\boldsymbol{\tilde{d}_{t}}$  which is used to generate words from vocabulary $P_{gen}(y_{t}|y_{<t},S)=softmax(\boldsymbol{W_{d}}\boldsymbol{\tilde{d}_{t}}+\boldsymbol{b_{d}})$
Also the conditional copy mechanism is adopted in base model. It introduces a variable $z_{t}$ to decide whether to copy from tables or generate from vocabulary. The probability to copy from table is $P(z_{t}=1|y_{<t}, S)=sigmoid(\boldsymbol{w_{e}} \cdot \boldsymbol{d_{t}}+b_{e})$.
Then it decomposes the conditional probability of generating $t^{th}$ word $P(y_{t}|y_{<t},S)$, given the tables $S$ and previously generated words $y_{<t}$, as follows.
\begin{equation}
    \label{cc-equ}
    \begin{split}
    &P(y_{t},z_{t}|y_{<t}, S)= \\
   &\left\{
\begin{array}{lll}
P(z_{t}=1|y_{<t},S)\sum_{y_{t}\leftarrow \boldsymbol{r_{i^{\prime}, j^{\prime}}}}\alpha_{t, i^{\prime}, j^{\prime}}       &      {z_{t}=1}\\
P(z_{t}=0|y_{<t},S)P_{gen}(y_{t}|y_{<t},S)     &      {z_{t} = 0}
\end{array} \right. 
\end{split}
\end{equation}

\section{Approach}

\label{approach-section}
In this section, we propose an effective hierarchical encoder to utilize three dimensional structure of input data in order to improve table representation. Those three dimensions include row, column and time. As shown in Figure \ref{modelstructure}, during encoding, our model consists of three layers including record encoders, record fusion gate and row-level encoder. Given tables $S$ as described in Section \ref{task-notations}, we first encode each record in each dimension respectively. Then we use the record fusion gate to combine them into a dense representation. Afterwards, we obtain row-level representation via mean pooling and self-attention with content selection gate. 
In decoding phase, the decoder can first find important row then attend to important record when generating texts. We describe model's details in following parts.

\subsection{Layer 1: Record Encoders}
\subsubsection{Row Dimension Encoder}
\label{row-encoder}
Based on our observation, when someone's points is mentioned in texts, some related records such as ``field goals made'' (FGM) and ``field goals attempted'' (FGA) will also be included in texts. Taken gold texts in Figure \ref{baseline-result} as example, when Al Jefferson's point 18 is mentioned, his FGM 9 and FGA 19 are also mentioned. Thus, when modeling a record, other records in the same row can be useful. 
Since the record in the row is not sequential, we use a self-attention network which is similar to \newcite{Liu2018LearningST} to model records in the context of other records in the same row. Let $\boldsymbol{r_{i, j}^{row}}$ be the row dimension representation of the record of $i^{th}$ row and $j^{th}$ column. 
Then, we obtain the context vector in row dimension $\boldsymbol{c_{i, j}^{row}}$ by attending to other records in the same row as follows. Please note that 
$\alpha_{i, j, j^{\prime}}^{row} \propto exp(\boldsymbol{r_{i, j}^{T}}\boldsymbol{W_{o}}\boldsymbol{{r}_{i, j^{\prime}}})$ is normalized across records in the same row $i$. $\boldsymbol{W_{o}}$ is a trainable parameter.
\begin{equation}
    \boldsymbol{c_{i, j}^{row}} = \sum_{j^{\prime},j^{\prime} \neq j}\alpha_{i, j, j^{\prime}}^{row}\boldsymbol{{r}_{i, j^{\prime}}}
    \label{row_context_vector}
\end{equation}

Then, we combine record's representation with $\boldsymbol{c_{i, j}}$ and obtain the row dimension record representation $\boldsymbol{r_{i, j}^{row}}=tanh(\boldsymbol{W_{f}}[\boldsymbol{r_{i, j}}; \boldsymbol{c_{i, j}^{row}}])$. $\boldsymbol{W_{f}}$ is a trainable parameter.

\subsubsection{Column Dimension Encoder}

Each input table consists of multiple rows and columns. Each column in the table covers one type of information such as points. Only few of the row may have high points or other type of information and thus become the important one. For example, in ``Column Dimension'' part of Figure \ref{modelstructure}, ``Al Jefferson'' is more important than ``Gary Neal'' because the former one have more impressive points. Therefore, when encoding a record, it is helpful to compare it with other records in the same column in order to understand the performance level reflected by the record among his teammates (rows).
We employ self-attention similar to the one used in Section \ref{row-encoder} in column dimension to compare between records. We let $\boldsymbol{{r_{i, j}^{col}}}$ be the column representation of the record of $i^{th}$ row and $j^{th}$ column. 
We obtain context vector in column dimension $\boldsymbol{c_{i, j}^{col}}$ as follows. Please note that $\alpha_{j, i, i^{\prime}}$ is normalized across records from different rows $i^{\prime}$ but of the same column $j$. The column dimension representation $\boldsymbol{r_{i, j}^{col}}$ is obtained similar to row dimension.
\begin{align}
    \boldsymbol{c_{i, j}^{col}} = \sum_{i^{\prime},i^{\prime} \neq i}\alpha_{j, i,i^{\prime}}^{col}\boldsymbol{{r}_{i^{\prime}, j}}
    \label{column_context_vector}
\end{align}
\subsubsection{Time Dimension Encoder}
As mentioned in Section \ref{intro}, we find some expressions in texts require information about players' historical (time dimension) performance. So the history information of record $\boldsymbol{r_{i,j}}$ is important. Note that we have already constructed timeline for each record entity and type as described in Section \ref{timeline-const}. Given those timelines, We collect records with same entity and type in the timeline which has date before date $k$ of the record $\boldsymbol{r_{i,j}}$ as history information. Since for some record, the history information can be too long, we set a history window. Thus, we keep most recent history information sequence within history window and denote them as $hist(\boldsymbol{r_{i, j}})$. We model this kind of information in time dimension via self-attention. However, unlike the unordered nature of rows and columns, the history information is sequential. Therefore, we introduce a trainable position embedding $\boldsymbol{emb_{pos}(k^{\prime})}$ and add it to the record's representation and obtain a new record representation $\boldsymbol{{rp}_{k^{\prime}}}$. It denotes the representation of a record with the same entity and type of $\boldsymbol{r_{i,j}}$ but of the date $k^{\prime}$ before $k$ in the corresponding history window.
We use $\boldsymbol{r_{i, j}^{time}}$ to denote the history representation of the record of $i^{th}$ row and $j^{th}$ column. Then the history dimension context vector is obtained by attending to history records in the window. Please note that we use 1-layer MLP as score function here and $\alpha_{k, k^{\prime}}^{time}$ is normalized within the history window. We obtain the time dimension representation $\boldsymbol{r_{i, j}^{time}}$ similar to row dimension.
\begin{align}
    \alpha_{k, k^{\prime}}^{time} \propto exp(score(\boldsymbol{{rp}_{k}}, \boldsymbol{{rp}_{k^{\prime}}}))
    \label{hist_attn_weight}\\
    \boldsymbol{c_{i, j}^{time}} = \sum_{k^{\prime}<k}\alpha_{k, k^{\prime}}^{time}\boldsymbol{{rp}_{k^{\prime}}}
    \label{hist_context_vector}
\end{align}

\subsection{Layer 2: Record Fusion Gate}

After obtaining record representations in three dimension, it is important to figure out which representation plays a more important role in reflecting the record's information. If a record stands out from other row's records of same column, the column dimension representation may have a higher weight in forming the overall record representation. If a record differs from previous match significantly, the history dimension representation may have a higher weight. Also, some types of information may appear in texts more frequently together which can be reflected by row dimension representation. Therefore, we propose a record fusion gate to adaptively combine all three dimension representations. First, we concatenate $\boldsymbol{r_{i, j}^{row}}$, $\boldsymbol{r_{i, j}^{col}}$ and $\boldsymbol{r_{i, j}^{time}}$, then adopt a 1-layer MLP to obtain a general representation $\boldsymbol{r_{i, j}^{gen}}$ which we consider as a baseline representation of records' information. 
Then, we compare each dimension representation with the baseline and obtain its weight in the final record representation. We use 1-layer MLP as the score function. Equation \ref{fusion-rep-weight} shows an example of calculating column dimension representation's weight in the final record representation. The weight of row and time dimension representation is obtained similar to the weight of column dimension representation. 
\begin{equation}
    \alpha_{fus}^{col} \propto exp(score(\boldsymbol{r_{i, j}^{col}}, \boldsymbol{r_{i, j}^{gen}}))
    \label{fusion-rep-weight}
\end{equation}
In the end, the fused record representation $\boldsymbol{\tilde{r}_{i, j}}$ is the weighted sum of the three dimension representations.
\begin{equation}
    \boldsymbol{\tilde{r}_{i, j}}=\alpha_{fus}^{row}\boldsymbol{r_{i, j}^{row}}+\alpha_{fus}^{col}\boldsymbol{r_{i, j}^{col}}+\alpha_{fus}^{time}\boldsymbol{r_{i, j}^{time}}
\end{equation}

\subsection{Layer 3: Row-level Encoder}
\label{row-level-encoder}
For each row, we combine its records via mean pooling (Equation \ref{row_rep}) in order to obtain a general representation of the row which may reflect the row (player or team)'s overall performance. $C$ denotes the number of columns.
\begin{equation}
    \boldsymbol{row_{i}} = MeanPooling({\boldsymbol{\tilde{r}_{i, 1}}}, \boldsymbol{\tilde{r}_{i, 2}}, ..., \boldsymbol{\tilde{r}_{i, C}})
    \label{row_rep}
\end{equation}
Then, we adopt content selection gate $\boldsymbol{g_{i}}$, which is proposed by \newcite{Puduppully} on rows' representations $\boldsymbol{row_{i}}$, and obtain a new representation $\boldsymbol{\tilde{row}_{i}}=\boldsymbol{g_{i}}\odot \boldsymbol{row_{i}}$ to choose more important information based on each row's context. 
\subsection{Decoder with Dual Attention}
\label{new-decoder}
Since record encoders with record fusion gate provide record-level representation and row-level encoder provides row-level representation. Inspired by \newcite{N18-2097}, we can modify the decoder in base model to first choose important row then attend to records when generating each word. Following notations in Section \ref{base-model}, 
$\beta_{t, i} \propto exp(score(\boldsymbol{d_{t}}, \boldsymbol{row_{i}}))$ obtains the attention weight with respect to each row. Please note that $\beta_{t, i}$ is normalized across all row-level representations from all three tables.
Then, $\gamma_{t, i, j} \propto exp(score(\boldsymbol{d_{t}}, \boldsymbol{\tilde{r}_{i, j}}))$ obtains attention weight for records. Please note that we normalize $\gamma_{t, i, j}$ among records in the same row.

We use the row-level attention $\beta_{t, i}$ as guidance for choosing row based on row's general representation. Then we use it to re-weight the record-level attention $\gamma_{t, i, j}$ and change the attention weight in base model to $\tilde{\alpha}_{t, i, j}$. Please note that ${\tilde{\alpha}_{t, i, j}}$ sum to 1 across all records in all tables.
\begin{equation}
    {\tilde{\alpha}_{t, i, j}} = \beta_{t, i}\gamma_{t, i, j}
    \label{dual-fused-attn}
\end{equation}

\subsection{Training}

Given a batch of input tables $\{S\}_{G}$ and reference output $\{Y\}_{G}$, we use negative log-likelihood as the loss function for our model. We train the model by minimizing $L$. $G$ is the number of examples in the batch and $T_{g}$ represents the length of $g^{th}$ reference's length.
\begin{equation}
L=-\frac{1}{G}\sum_{g=1}^{G}\sum_{t=1}^{T_{g}}logP(y_{t,g}|y_{<t,g}, S_{g}) \label{full-loss-func}
\end{equation}

\begin{table*}[ht]
	
	\begin{center}
	    \begin{tabular}{p{6.1cm} p{0.9cm}<{\centering}p{0.9cm}<{\centering}p{0.9cm}<{\centering}  p{0.9cm}<{\centering}p{0.9cm}<{\centering}p{0.9cm}<{\centering} p{0.9cm}<{\centering} }
	        \hline
			& \multicolumn{7}{c}{\bf  Development}\\
			\hline
			\centering
			\multirow{1}{*}[ - 0.2cm]{\bf Model} & \multicolumn{2}{c}{\bf RG} & \multicolumn{3}{c}{\bf CS}& \multicolumn{1}{c}{\bf CO}& \multirow{1}{*}[ - 0.2cm]{\bf BLEU}\\
			\bf & \bf P\% & \bf \# &\bf P\% &\bf R\% &\bf F1\% &\bf DLD\% & \\
	    
			\hline
			Gold &  94.79 & 23.31 & 100.00 & 100.00 & 100.00 & 100.00 & 100.00 \\ 
			Template & \textbf{99.92} & \textbf{54.23} & 26.60 & \textbf{59.13} & 36.69 & 14.39 & 8.62 \\
            CC \cite{Wiseman} & 75.10 & 23.95 & 28.11 & 35.86 & 31.52 & 15.33 & 14.57 \\
            NCP+CC \cite{Puduppully} & 87.51 & 33.88 & 33.52 & 51.21 & 40.52 & 18.57 & 16.19 \\
            \hline
            Hierarchical LSTM Encoder & 91.59 & 32.56 & 31.62 & 44.22 & 36.87 & 17.49 & 15.21 \\
            Hierarchical CNN Encoder & 90.86 & 30.59 & 30.32 & 40.28 & 34.60 & 15.75  & 14.08 \\
            Hierarchical SA Encoder & 90.46 & 29.82 & 34.39 & 45.43 & 39.15 & 19.81 & 15.62 \\
            Hierarchical MHSA Encoder & 92.87 & 28.42 & 34.87 & 42.41 & 38.27 & 18.28 & 15.12 \\
            \hline
            CC (Our implementation) & 76.50 & 22.48 & 29.18 & 34.22 & 31.50 & 15.43 & 13.65 \\
            Our Model & 91.84 & 32.11 & \textbf{35.39} & 48.98 & \textbf{41.09} & \textbf{20.70} & \textbf{16.24} \\
            \quad \emph{-row-level encoder} & 90.19 & 27.90 & 34.70 & 42.53 & 38.22 & 20.02 & 15.32 \\
            \quad \emph{-row} & 91.08 & 30.95 & 35.03 & 47.09 & 40.17 & 20.03 & 15.50 \\
            \quad \emph{-column} & 91.66 & 28.63 & 34.83 & 43.62 & 38.73 & 19.59 & 15.99 \\
            \quad \emph{-time} & 90.94 & 31.43 & 34.62 & 47.74 & 40.13 & 19.81 & 16.10 \\
            \quad \emph{-position embedding} & 89.97 & 28.37 & 34.72 & 43.69 & 38.69 & 19.54 & 16.05 \\
            \quad \emph{-record fusion gate} & 89.34 & 32.22 & 32.28 & 46.68 & 38.17 & 18.49 & 14.97 \\

            \hline
			 & \multicolumn{7}{c}{\bf  Test}\\
			 \hline
			Gold & 94.89 & 24.14  & 100.00 & 100.00  & 100.00  & 100.00  & 100.00  \\ 
			Template & \bf 99.94  & \bf 54.21 & 27.02 & \bf 58.22 &  36.91 & 15.07  & 8.58 \\

            CC \cite{Wiseman} & 74.80 & 23.72 & 29.49 & 36.18 & 32.49 & 15.42 & 14.19\\
            OpAtt \cite{Nie} & - & -& - & -& -& -& 14.74\\
            NCP+CC \cite{Puduppully} &  87.47 & 34.28 & 34.18 &  51.22 & 41.00  & 18.58  & 16.50 \\
            CC (Our implementation) & 75.37 & 22.32 & 28.91 & 33.12 & 30.87 & 15.34 & 14.02 \\
            Our model & 91.46 & 31.47 & \bf 36.09 & 48.01 & \bf 41.21 & \bf 20.86 &  \bf 16.85 \\
			\hline
		\end{tabular}
        \end{center}
    \vspace{-0.2cm}
    \caption{Automatic evaluation results. Results were obtained using \newcite{Puduppully}'s updated models}
    \vspace{-0.4cm}
	\label{exp-full-result}
\end{table*}

\section{Experiments}

\subsection{Dataset and Evaluation Metrics}

We conducted experiments on ROTOWIRE \cite{Wiseman}. For each example, it provides three tables as described in Section \ref{task-notations} which consists of 628 records in total with a long game summary. The average length of game summary is 337.1. In this paper, we followed the data split introduced in \newcite{Wiseman}: 3398 examples in training set, 727 examples in development set and 728 examples in test set.
We followed \newcite{Wiseman}'s work and use BLEU \cite{Papineni2002BleuAM} and three extractive evaluation metrics RG, CS and CO \cite{Wiseman} for evaluation. 
The main idea of the extractive evaluation metrics is to use an Information Extraction (IE) model to identify records mentioned in texts. Then compare them with tables or records extracted from reference to evaluate the model.
RG (Relation Generation) measures content fidelity of texts.
CS (Content Selection) measures model's ability on content selection. 
CO (Content Ordering) measures model's ability on ordering the chosen records in texts. We refer the readers to \newcite{Wiseman}'s paper for more details.

\subsection{Implementation Details}

Following configurations in \newcite{Puduppully}, we set word embedding and LSTM decoder hidden size as 600. The decoder's layer was set to be 2. Input feeding \cite{D15-1166} was also used for decoder. We applied dropout at a rate 0.3. For training, we used Adagrad \cite{Duchi2010AdaptiveSM} optimizer with learning rate of 0.15, truncated BPTT (block length 100), batch size of 5 and learning rate decay of 0.97. For inferring, we set beam size as 5.
We also set the history windows size as 3 from \{3,5,7\} based on the results. Code of our model can be found at \href{https://github.com/ernestgong/data2text-three-dimensions/}{https://github.com/ernestgong/data2text-three-dimensions/}.

\subsection{Results}

\subsubsection{Automatic Evaluation}

Table \ref{exp-full-result} displays the automatic evaluation results on both development and test set. We chose Conditional Copy (CC) model as our baseline, which is the best model in \newcite{Wiseman}. We included reported scores with updated IE model by \newcite{Puduppully} and our implementation's result on CC in this paper.
Also, we compared our models with other existing works on this dataset including OpATT \cite{Nie} and Neural Content Planning with conditional copy (NCP+CC) \cite{Puduppully}. 
In addition, we implemented three other hierarchical encoders that encoded tables' row dimension information in both record-level and row-level to compare with the hierarchical structure of encoder in our model. The decoder was equipped with dual attention \cite{N18-2097}. The one with LSTM cell is similar to the one in \newcite{N18-2097} with 1 layer from \{1,2,3\}. The one with CNN cell \cite{Gehring2017ConvolutionalST} has kernel width 3 from \{3, 5\} and 10 layer from \{5,10,15,20\}. The one with transformer-style encoder (MHSA)  \cite{Vaswani2017AttentionIA} has 8 head from \{8, 10\} and 5 layer from \{2,3,4,5,6\}. 
The heads and layers mentioned above were for both record-level encoder and row-level encoder respectively. The self-attention (SA) cell we used, as described in Section \ref{approach-section}, achieved better overall performance in terms of F1\% of CS, CO and BLEU among the hierarchical encoders.
Also we implemented a template system same as the one used in \newcite{Wiseman} which outputted eight sentences: an introductory sentence (two teams' points and who win), six top players' statistics (ranked by their points) and a conclusion sentence. We refer the readers to \newcite{Wiseman}'s paper for more detailed information on templates. The gold reference's result is also included in Table \ref{exp-full-result}.
Overall, our model performs better than other neural models on both development and test set in terms of RG's P\%, F1\% score of CS, CO and BLEU, indicating our model's clear improvement on generating high-fidelity, informative and fluent texts.
Also, our model with three dimension representations outperforms hierarchical encoders with only row dimension representation on development set. This indicates that cell and time dimension representation are important in representing the tables.
Compared to reported baseline result in \newcite{Wiseman}, we achieved improvement of $22.27\%$ in terms of RG, $26.84\%$ in terms of CS F1\%, $35.28\%$ in terms of CO and $18.75\%$ in terms of BLEU on test set. 
Unsurprisingly, template system achieves best on RG P\% and CS R\% due to the included domain knowledge. Also, the high RG \# and low CS P\% indicates that template will include vast information while many of them are deemed redundant. In addition, the low CO and low BLEU indicates that the rigid structure of the template will produce texts that aren't as adaptive to the given tables and natural as those produced by neural models.
Also, we conducted ablation study on our model to evaluate each component's contribution on development set. Based on the results, the absence of row-level encoder hurts our model's performance across all metrics especially the content selection ability. 

\begin{table}[h]
	\begin{center}
	    \begin{tabular}{p{0.69cm} p{0.48cm}<{\centering}p{0.48cm}<{\centering}p{0.68cm}<{\centering}  p{0.68cm}<{\centering}p{0.72cm}<{\centering}p{0.68cm}<{\centering} }
	        \hline
			\centering
			\multirow{2}{*}{\bf Model} & \multicolumn{2}{c}{\bf RG} & \multicolumn{2}{c}{\bf CS}& \multicolumn{1}{c}{\bf CO}& \multirow{2}{*}{\bf BLEU}\\
			& \bf P\% & \bf \# &\bf P\% &\bf R\% &\bf DLD\% & \\
			 \hline
			Gold & 96.01 & 17.17 & 100.00  & 100.00 & 100.00  & 100.00  \\ 
			TEM & \bf 99.97  & \bf 54.14 & 23.88 & \bf 72.63  & 11.90 & 8.33 \\
            CC & 75.26 & 16.37 & 32.63  & 39.62 & 15.34 & 14.03 \\
            DEL${}^{*}$ & 84.86 & 19.31 & 30.81 & 38.79 & 16.34 & 16.19 \\
            NCP & 87.99  & 24.50 & 35.97 & 55.85  & 16.98  & 16.22 \\
            Ours & 92.51 & 22.73 & \bf38.52  & 52.98 & \bf19.95  &  \bf 16.69\\
			\hline
		\end{tabular}
        \end{center}
    \caption{Automatic evaluation results on test set. Results were obtained using \newcite{Wiseman}'s trained extractive evaluation models with relexicalization \cite{Li}. ${}^{*}$ We include delayed copy (DEL)'s result in the paper \cite{Li} for comparison.}
    \vspace{-0.3cm}
	\label{exp-extended-result}
\end{table}

Row, column and time dimension information are important to the modeling of tables because subtracting any of them will result in performance drop. Also, position embedding is critical when modeling time dimension information according to the results. In addition, record fusion gate plays an important role because BLEU, CO, RG P\% and CS P\% drop significantly after subtracting it from full model.
Results show that each component in the model contributes to the overall performance.
In addition, we compare our model with delayed copy model (DEL) \cite{Li} along with gold text, template system (TEM), conditional copy (CC) \cite{Wiseman} and NCP+CC (NCP) \cite{Puduppully}. \newcite{Li}'s model  generate a template at first and then fill in the slots with delayed copy mechanism. Since its result in \newcite{Li}'s paper was evaluated by IE model trained by \newcite{Wiseman} and ``relexicalization'' by \newcite{Li}, we adopted the corresponding IE model and re-implement ``relexicalization'' as suggested by \newcite{Li} for fair comparison. Please note that CC's evaluation results via our re-implemented ``relexicalization'' is comparable to the reported result in \newcite{Li}. We applied them on models other than DEL as shown in Table \ref{exp-extended-result} and report DEL's result from \cite{Li}'s paper. It shows that our model outperform \newcite{Li}'s model significantly across all automatic evaluation metrics in Table \ref{exp-extended-result}.

\subsubsection{Human Evaluation}

\begin{table}[h]
	\begin{center}
 	\begin{tabular}
    {p{0.7cm}>{\raggedleft}p{0.7cm}>{\raggedleft}p{0.7cm}>{\raggedleft}p{1.0cm}>{\raggedleft}p{1.0cm}>{\raggedleft\arraybackslash}p{1.0cm}}
			\hline  \bf Model & \bf \#Sup &  \bf \#Cont & \bf \#Gram &  \bf \#Coher &  \bf \#Conc \\
            \hline
            Gold & 3.48 & 0.19 & 16.67 & 24.22 & 25.78 \\
            Temp & 7.83 & 0.00 & 11.56 & -16.67 & 21.11 \\
            CC & 3.91 & 1.23 & -11.33 & -7.78 & -28.00 \\
            NCP & 5.15 & 0.82 & -17.33 & -5.33 & -17.11 \\
            Ours & 3.63 & 0.44  & 0.44 & 5.56 & -1.78 \\
            \hline
			
		\end{tabular}
	\end{center}
    \caption{Human evaluation results.}
    \vspace{-0.6cm}
	\label{human-eval-2}
\end{table}
In this section, we hired three graduates who passed intermediate English test (College English Test Band 6) and were familiar with NBA games to perform human evaluation.

First, in order to check if history information is important, we sampled 100 summaries from training set and asked raters to manually check whether the summary contained expressions that need to be inferred from history information. It turns out that $56.7\%$ summaries of the sampled summaries need history information.

Following human evaluation settings in \newcite{Puduppully}, we conducted the following human evaluation experiments at the same scale. The second experiment is to assess whether the improvement on relation generation metric reported in automatic evaluation is supported by human evaluation. 
We compared our full model with gold texts, template-based system, CC \cite{Wiseman} and NCP+CC (NCP) \cite{Puduppully}. 
We randomly sampled 30 examples from test set.
Then, we randomly sampled 4 sentences from each model's output for each example.
We provided the raters of those sampled sentences with the corresponding NBA game statistics. They were asked to count the number of supporting and contradicting facts in each sentence. Each sentence is rated independently. We report the average number of supporting facts (\#Sup) and contradicting facts (\#Cont) in Table \ref{human-eval-2}. Unsurprisingly, template-based system includes most supporting facts and least contradicting facts in its texts because the template consists of a large number of facts and all of those facts are extracted from the table. Also, our model produces less contradicting facts than other two neural models. Although our model produces less supporting facts than NCP and CC, it still includes enough supporting facts (slightly more than gold texts). Also, comparing to NCP+CC (NCP)’s tendency to include vast information that contain redundant information, our model’s ability to select and accurately convey information is better. All other results (Gold, CC, NCP and ours) are significantly different from template-based system's results in terms of number of supporting facts according to one-way ANOVA with posthoc Tukey HSD tests. All significance difference reported in this paper are less than 0.05. Our model is also significantly different from the NCP model. As for average number of contradicting facts, our model is significantly different from other two neural models. Surprisingly, gold texts were found containing contradicting facts. We checked the raters's result and found that gold texts occasionally include wrong field-goal or three-point percent or wrong points difference between the winner and the defeated team. We can treat the average contradicting facts number of gold texts as a lower bound.

In the third experiment, following \newcite{Puduppully}, we asked raters to evaluate those models in terms of grammaticality (is it more fluent and grammatical?), coherence (is it easier to read or follows more natural ordering of facts? ) and conciseness (does it avoid redundant information and repetitions?). We adopted the same 30 examples from above and arranged every 5-tuple of summaries into 10 pairs. Then, we asked the raters to choose which system performs the best given each pair. Scores are computed as the difference between percentage of times when the model is chosen as the best and percentage of times when the model is chosen as the worst. 
Gold texts is significantly more grammatical than others across all three metrics. Also, our model performs significantly better than other two neural models (CC, NCP) in all three metrics. Template-based system generates significantly more grammatical and concise but significantly less coherent results, compared to all three neural models. Because the rigid structure of texts ensures the correct grammaticality and no repetition in template-based system's output. However, since the templates are stilted and lack variability compared to others, it was deemed less coherent than the others by the raters.

\subsubsection{Qualitative Example}

\begin{figure}[h]
   \centering
   \begin{center}
     \includegraphics*[width=1.0\linewidth]{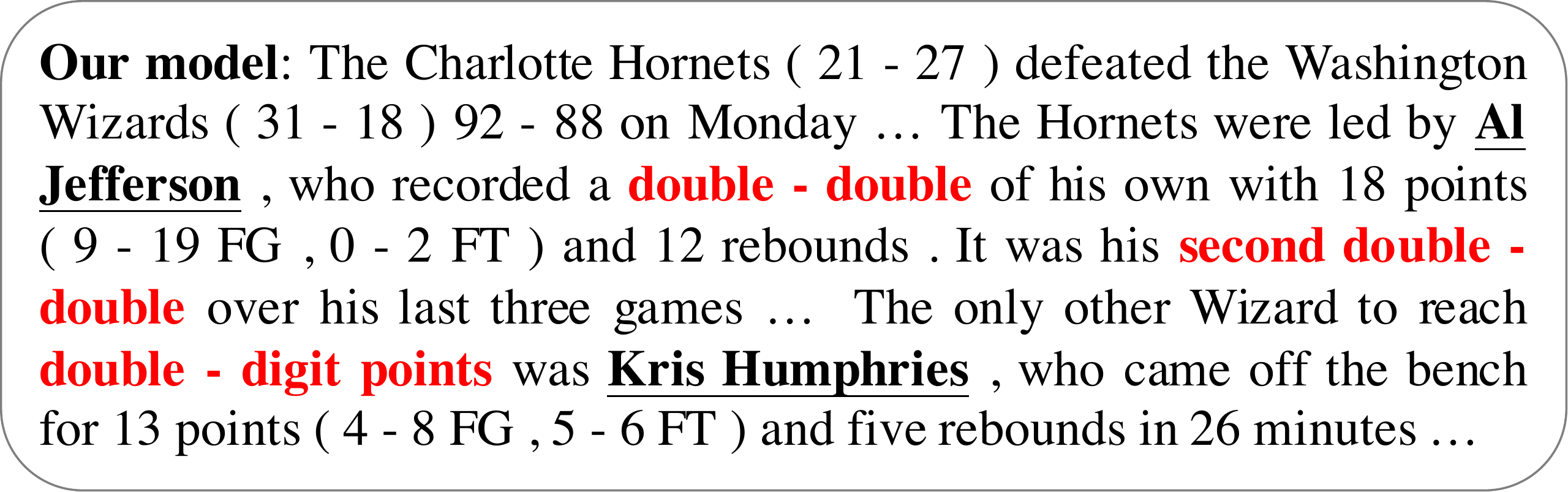}
   \caption{ An generation example of our model based on the same tables in Figure 1. Text that accurately reflects players (Al Jefferson and Kris Humphries) performance is in red.}
   \label{example}
   \vspace{-0.3cm}
   \end{center}
\end{figure}

Figure \ref{example} shows an example generated by our model.
It evidently has several nice properties:
it can accurately select important player ``Al Jefferson'' from the tables who is neglected by baseline model, which need the model to understand performance difference of a type of data (column) between each rows (players). Also it correctly summarize performance of ``Al Jefferson'' in this match as ``double-double'' which requires ability to capture dependency from different columns (different type of record) in the same row (player). In addition, it models ``Al Jefferson'' history performance and correctly states that ``It was his second double-double over his last three games'', which is also mentioned in gold texts included in Figure 1 in a similar way.  

 \section{Related Work}

In recent years, neural data-to-text systems make remarkable progress on generating texts directly from data. 
\newcite{Mei2016WhatTT} proposes an encoder-aligner-decoder model to generate weather forecast, while \newcite{N18-2098} propose a mixed hierarchical attention. \newcite{AAAI1816203} proposes a hybrid content- and linkage-based attention mechanism to model the order of content.
\newcite{AAAI1816599} propose to integrate field information into table representation and enhance decoder with dual attention.
\newcite{AAAI1816138} develops a table-aware encoder-decoder model.
\newcite{Wiseman} introduced a document-scale data-to-text dataset, consisting of long text with more redundant records, which requires the model to select important information to generate. We describe recent works in Section \ref{intro}.
Also, some studies in abstractive text summarization encode long texts in a hierarchical manner. \newcite{N18-2097} uses a hierarchical encoder to encode input, paired with a discourse-aware decoder. \newcite{W17-4505} encode document hierarchically and propose coarse-to-fine attention for decoder. Recently, \newcite{liu2019hierarchical} propose a hierarchical encoder for data-to-text generation which uses LSTM as its cell. \newcite{murakami-etal-2017-learning} propose to model stock market time-series data and generate comments.
As for incorporating historical background in generation, \newcite{robin1994revision} proposed to build a draft with essential new facts at first, then incorporate background facts when revising the draft based on functional unification grammars. Different from that, we encode the historical (time dimension) information in the neural data-to-text model in an end-to-end fashion. Existing works on data-to-text generation neglect the joint representation of tables' row, column and time dimension information. In this paper, we propose an effective hierarchical encoder which models information from row, column and time dimension simultaneously.

\section{Conclusion}

In this work, we present an effective hierarchical encoder for table-to-text generation that learns table representations from row, column and time dimension.
In detail, our model consists of three layers, which learn records' representation in three dimension, combine those representations via their sailency and obtain row-level representation based on records' representation. Then, during decoding, it will select important table row before attending to records.
Experiments are conducted on ROTOWIRE, a benchmark dataset of NBA games. 
Both automatic and human evaluation results show that our model achieves the new state-of-the-art performance.

\section*{Acknowledgements} We would like to thank the anonymous reviewers for their helpful comments. We'd also like to thank Xinwei Geng, Yibo Sun, Zhengpeng Xiang and Yuyu Chen for their valuable input. This work was supported by the National Key R\&D Program of China via grant 2018YFB1005103 and National Natural Science Foundation of China (NSFC) via grant 61632011 and 61772156.

\bibliography{emnlp-ijcnlp-2019}
\bibliographystyle{acl_natbib}

\end{document}